%% file: main.tex
\documentclass[10pt,twocolumn,letterpaper]{article}

\providecommand{\wacvopts}{algorithms,pagenumbers}
\expandafter\usepackage\expandafter[\wacvopts]{wacv}

\input{preamble}

\definecolor{wacvblue}{rgb}{0.21,0.49,0.74}
\ifdefined\wacvsplit\usepackage{xr-hyper}\fi
\usepackage[pagebackref,breaklinks,colorlinks,allcolors=wacvblue]{hyperref}
\ifdefined\wacvsplit
\makeatletter
\def\wacv@unlink#1{\@ifundefined{r@#1}{}{%
  \expandafter\xdef\csname r@#1\endcsname{{%
  \expandafter\expandafter\expandafter\@car\csname r@#1\endcsname\@nil}{}{}{}{}}}}
\AtBeginDocument{\wacv@unlink{sec:supp:kernels}\wacv@unlink{sec:supp:transfer}}
\makeatother
\fi

\def\wacvPaperID{2130}
\def\confName{WACV}
\def\confYear{2027}

\title{GPA-VGGT: Geometry-Preserving Acceleration of VGGT}

\author{Muyeol Sung \qquad Jangwon Kim \qquad Jongwoo Lim
\\[2pt]
Seoul National University\\
{\tt\small \{smy4024169, jangwons9, jongwoo.lim\}@snu.ac.kr}%
\iftoggle{wacvfinal}{\\[6pt]}{}%
}

\renewcommand{\arraystretch}{0.96}

\begin{document}
\maketitle

\nocite{vggt,
        fastvggt,avggt,litevggt,flashvggt,fastervggt,quantvggt,qvggt,come,httm,
        dust3r,fast3r,pi3,vggtomega,
        tome,tomesd,
        co3d,re10k,sevenscenes,nrgbd,
        flashattention,triton,pytorch}

\input{sec/0_abstract}
\input{sec/1_intro}
\input{sec/2_related}
\input{sec/4_analysis}

\input{sec/5_method}

\input{sec/6_experiments}

\input{sec/8_conclusion}

\clearpage
{
    \small
    \bibliographystyle{ieeenat_fullname_order}
    \bibliography{main}
}

\ifdefined\wacvsplit\else\input{supp/0_supp}\fi %

\end{document}

%% file: preamble.tex
\usepackage{lmodern}

\usepackage{multirow}
\usepackage{needspace}

\usepackage{tikz}
\usetikzlibrary{positioning, fit, calc, arrows.meta, decorations.pathreplacing,
                backgrounds, shapes.geometric}
\usepackage{graphicx}

\usepackage{array}

\definecolor{plancol}{rgb}{0.16,0.38,0.60}
\definecolor{fillcol}{rgb}{0.945,0.955,0.970}
\definecolor{tabcol}{rgb}{0.97,0.955,0.925}
\definecolor{figcol}{rgb}{0.93,0.955,0.935}

\newcommand{\rev}[1]{{\color{red}#1}}
\renewcommand{\rev}[1]{#1}   %

\newcommand{\new}[1]{{\color{red}#1}}
\renewcommand{\new}[1]{#1}   %

\newcommand{\crr}[1]{{\color{red}#1}}
\iftoggle{wacvfinal}{\renewcommand{\crr}[1]{#1}}{}
\AtBeginDocument{\ifdefined\pdfstringdefDisableCommands
  \pdfstringdefDisableCommands{\def\crr#1{#1}\def\rev#1{#1}}\fi}

%% file: sec/0_abstract.tex
\begin{abstract}
Feed-forward geometry models such as VGGT recover camera parameters, depth maps,
and point maps from one to hundreds of views in a single forward pass, making them
powerful backbones for multi-view 3D perception. However, VGGT's global attention is
quadratic in the number of frames and dominates \rev{end-to-end runtime} on long sequences, which
has \rev{motivated a range of} training-free efficiency methods such as token
merging. These methods, however, come with a drop in accuracy that makes them hard
to use in place of VGGT. We propose GPA-VGGT, a training-free acceleration method
\rev{that replaces global attention in early blocks with frame attention augmented by a
shared cross-frame K/V set. In the remaining blocks, we merge only keys and values
while preserving all queries, restore the attention mass of the tokens each merged
key represents, and randomly project the matching metric to 128 dimensions to reduce matching
cost. Across diverse multi-view geometry tasks, GPA-VGGT provides training-free
acceleration at speedups comparable to prior work while preserving the information
required for precise geometric estimation, retaining over $99\%$ of VGGT's pose
accuracy under AUC@3.}
\iftoggle{wacvfinal}{\crr{Our project page is available at
\url{https://noheat61.github.io/GPA-VGGT/}.}}{}
\end{abstract}

%% file: sec/1_intro.tex
\section{Introduction}
\label{sec:intro}

After pairwise systems such as DUSt3R~\cite{dust3r}, feed-forward models including
Fast3R~\cite{fast3r}, $\pi^3$~\cite{pi3}, and VGGT~\cite{vggt} extended direct 3D reconstruction to multi-view inputs and have
become widely used backbones for multi-view 3D perception. VGGT jointly predicts
camera parameters, depth maps, point maps, and tracks from one to hundreds of
unposed views. However, this accuracy comes at a substantial computational cost:
its aggregator alternates between \rev{frame} attention, whose complexity scales
linearly with the number of views, and global attention, whose complexity scales
quadratically. On long sequences, global attention dominates \rev{end-to-end runtime} and
becomes the primary bottleneck, limiting the scalability of VGGT to the many-view
settings it was designed to support.

\input{fig/1_teaser}

This bottleneck has motivated a growing body of acceleration methods.
FastVGGT~\cite{fastvggt} introduces \crr{Token Merging (ToMe)~\cite{tome}} into the
aggregator, whereas FasterVGGT~\cite{fastervggt} replaces dense global attention with
block-sparse attention. AVGGT~\cite{avggt} replaces early global-attention blocks with
\rev{frame} attention and subsamples keys in later blocks, while
LiteVGGT~\cite{litevggt} uses geometry-aware cached token merging and
FlashVGGT~\cite{flashvggt} compresses the global aggregator with learned descriptors,
both \crr{relying} on fine-tuning for acceleration.
Although these methods report substantial speedups with little loss in accuracy,
evaluation under a common protocol shows that this accuracy preservation is
largely confined to lenient metrics. At relaxed pose thresholds \rev{(\eg, AUC@$30$)}, the
degradation may appear minor; however, as Figure~\ref{fig:teaser} shows, camera estimates
become noticeably less precise under strict thresholds \rev{(\eg, AUC@$3$)}. \rev{The former primarily reflects coarse
pose correctness and can mask errors that remain consequential for reconstruction,
whereas the latter directly tests the geometric precision required for an accelerated
method to serve as a practical replacement for VGGT.}

We introduce GPA-VGGT, a training-free acceleration method designed to preserve
the information required for accurate geometric estimation. GPA-VGGT merges keys
and values only, retaining every query so that each token continues to produce its
own output. It further adds $\log s$ to the logit of each merged or pooled key,
\rev{restoring the attention mass of the tokens it represents}. In early blocks,
GPA-VGGT preserves cross-frame context through a compact \rev{shared
cross-frame K/V set} comprising the camera and register tokens from all frames, together with a
small number of content-adaptive pooled patch summaries. Finally, because the
retained similarity is used only for an argmax operation, we reduce matching cost
through a fixed low-dimensional projection of the matching metric.

Without additional training, GPA-VGGT processes a 200-frame sequence in
approximately $34$\,s, compared with $87$\,s for vanilla VGGT, while matching the
runtime of other baselines. Across all benchmarks evaluated using AUC@3, GPA-VGGT remains within $1\%$ of vanilla VGGT on benchmarks and shows a particularly large advantage over existing acceleration methods on the more challenging setting. 

Our contributions are summarized below:
\begin{itemize}
  \item \rev{We analyze accuracy degradation in existing VGGT acceleration methods,
  including FastVGGT and AVGGT, and identify three distinct sources: query
  merging, attention-mass distortion, and early global to frame attention replacement.}
  \item We combine Pooled Cross-Frame Broadcast to retain early \rev{cross-frame}
  context with Query-Preserving K/V Merging to compress later global attention
  \rev{while retaining every query token}, and use low-dimensional matching to reduce
  overhead.
  \item We evaluate GPA-VGGT on representative benchmarks, maintaining
strong performance across multi-view geometry tasks without fine-tuning.
\end{itemize}

\pagebreak

%% file: fig/1_teaser.tex
\begin{figure}[t]
  \centering
  \includegraphics[width=\linewidth]{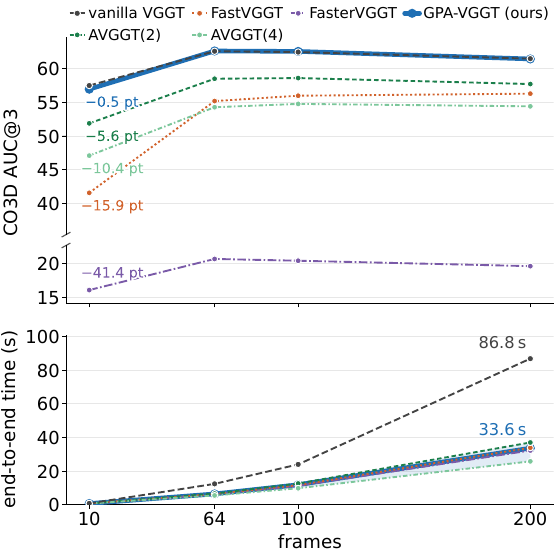}
  \caption{\textbf{\rev{Accuracy and runtime scaling with the number of input
  frames.}} Although existing \rev{acceleration methods} substantially reduce end-to-end
  \rev{runtime}, they incur pronounced camera-pose degradation under the strict
  AUC@3 metric. GPA-VGGT achieves comparable runtime while maintaining
  camera-pose accuracy close to that of vanilla VGGT.}
  \label{fig:teaser}
\end{figure}

%% file: sec/2_related.tex
\section{Related Work}
\label{sec:related}

\subsection{VGGT Acceleration}

VGGT~\cite{vggt} is a feed-forward multi-view geometry model that takes a set of unposed images as
input and jointly predicts camera parameters, depth maps, point maps, and tracking
features for all views. \rev{Each view is tokenized by a DINOv2~\cite{dinov2} encoder,
and its} Transformer~\cite{transformer} alternates between \rev{frame} attention and
global attention. Although global attention is essential for establishing multi-view
consistency, its quadratic complexity with respect to the number of tokens makes it the
primary computational bottleneck for long image sequences.

FlashVGGT~\cite{flashvggt} compresses the global aggregator using learned descriptors.
LiteVGGT~\cite{litevggt} introduces geometry-aware cached token merging, preserving
geometrically important tokens while reusing merge indices across adjacent blocks.
\rev{Co-Me~\cite{come} uses an auxiliary confidence predictor, trained by
distillation, to identify tokens suitable for merging.}
QuantVGGT~\cite{quantvggt} and QVGGT~\cite{qvggt} improve computational and memory
efficiency through reduced-precision inference. Because these methods rely on
\rev{additional training} or reduced precision, they are not the target of our
training-free architectural design.

FastVGGT~\cite{fastvggt} was the first method to introduce token merging into VGGT. To
address token collapse in long sequences, it uses a partitioning strategy tailored to the
multi-frame layout and preserves a subset of salient tokens from merging.
FasterVGGT~\cite{fastervggt} replaces dense global attention with training-free
block-sparse attention, retaining selected block interactions based on coarse attention
patterns and geometrically meaningful correspondences. AVGGT~\cite{avggt} replaces early
global blocks with \rev{frame} attention and applies uniform grid-based subsampling to
patch tokens in the remaining global blocks. Its design is motivated by the observation
that global attention supports cross-view alignment through sparse spatial
correspondences rather than dense token-to-token interactions.
\rev{HTTM~\cite{httm} performs head-wise temporal token merging with temporal reordering
to preserve feature diversity after head concatenation.}
Although these methods
provide substantial acceleration, their accuracy degradation becomes more apparent under
strict metrics such as AUC@3, which are sensitive to fine-grained camera-pose errors. Our
method instead preserves the information required for precise geometric estimation and pose accuracy at AUC@3.

\iftoggle{wacvfinal}{}{\newpage}
\subsection{Token Merging}

\crr{ToMe}~\cite{tome} was first introduced as a training-free approach to improve the
throughput of vision transformers, and was later extended to diffusion, video, and language models.
ToMe partitions tokens in each block into source and destination sets and merges tokens
based on the similarity of their attention keys. Since a merged key represents multiple
tokens, ToMe adds the logarithm of the group size to the attention logit \rev{so that the
merged key receives their combined attention mass}. ToMeSD~\cite{tomesd} adapts this procedure to Stable Diffusion by
selecting destination tokens from a fixed 2D grid and computing similarity from the
block input rather than attention keys. It also omits \rev{this} correction and
uses a merge--unmerge scheme to preserve dense outputs. FastVGGT adopts this
ToMeSD-style token merging for a 3D geometry backbone, where such design choices can
lead to fine-grained information loss and accuracy degradation in precise geometry
estimation.

%% file: sec/4_analysis.tex
\section{Analysis}
\label{sec:whydrop}

\subsection{\rev{Token Merging Disrupts Token-Specific Predictions and Attention Mass}}
\label{sec:whydrop:fast}

FastVGGT~\cite{fastvggt} merges queries together with keys and values, then restores the original
sequence length by assigning each merged output to all tokens in its group.
Consequently, non-representative tokens receive outputs that were not computed from
their individual queries. This differs fundamentally from an imperfect approximation
that can be corrected after merging. FastVGGT preserves salient tokens from merging, but
this only partially mitigates the loss of token-specific predictions.
Figure~\ref{fig:query} (top) plots the member-token outputs associated with twelve
representative merge slots selected from block 12, projected onto the first two
principal components of the unmerged output field. Query merging collapses within-group
output diversity to zero, whereas K/V-only merging closely overlaps with the unmerged outputs in this diagnostic.

\input{fig/2_query}

\rev{The number of original tokens represented by each merged key, its token size
$s$, is accounted for by adding $+\log s$ to the corresponding attention logit.} Following ToMeSD~\cite{tomesd}, FastVGGT omits
this correction, so a merged key receives the mass of only a single token. In VGGT,
this omission introduces an asymmetric bias because the first frame consists
predominantly of destination tokens; attention is therefore progressively biased toward
the first frame across merged blocks. Figure~\ref{fig:query} (bottom) shows that,
without the correction, the first frame receives nearly twice as much attention mass as
in the unmerged model, whereas applying the correction restores a distribution close to
the unmerged one. This effect causes a larger accuracy loss than query merging, and the
losses from the two factors are approximately additive.

\subsection{\crr{Early Blocks Still Need Cross-Frame Context}}
\label{sec:whydrop:avggt}

AVGGT's layer-wise analysis~\cite{avggt} concludes that early global-attention blocks do
not form
meaningful cross-view correspondences, motivating their conversion to \rev{frame}
attention. This finding supports replacing dense global attention, but does not
establish that early blocks can dispense with cross-frame context entirely. The absence
of token-level correspondences does not preclude informative interactions at the
regional level, where compact cross-frame context can still support geometric
estimation.

Figure~\ref{fig:crossframe} demonstrates that the principal limitation of early
\rev{frame} attention is not merely representational drift, but the disruption of a
shared geometric reference across views. In the top row of Figure~\ref{fig:crossframe}, \rev{frame}
attention displaces the block-8 output field by $15.5\%$ of its radius relative to the
full-global-attention reference, whereas the \rev{shared cross-frame K/V set} reduces this
displacement to $7.8\%$. The bottom row further reveals that this drift
is view-specific rather than globally shared: the pairwise cosine similarities between
\rev{per-frame} drift directions exhibit a broad distribution, with some views moving
coherently and others in opposing directions. Their mean similarity is \(-0.09\), close to zero, indicating the absence
of a common drift direction. Whereas a common drift preserves relative geometry, such
inconsistent view-specific perturbations alter inter-view relationships and degrade
relative-pose estimation. Adding shared special tokens and pooled patch summaries
progressively aligns the \rev{per-frame} drifts, thereby preserving the shared geometric
reference.

Early blocks require cross-frame context because \rev{frame} attention induces
view-specific representational drift, disrupting the relative geometry required for pose
estimation. Preserving this shared geometric reference, however, does not require the
full cross-frame key set; a compact shared context is sufficient. Incorporating
cross-frame special tokens partially mitigates the drift, and augmenting them with
image-token summaries yields further improvement.

%% file: fig/2_query.tex
\begin{figure}[t]
  \centering
  \includegraphics[width=\linewidth]{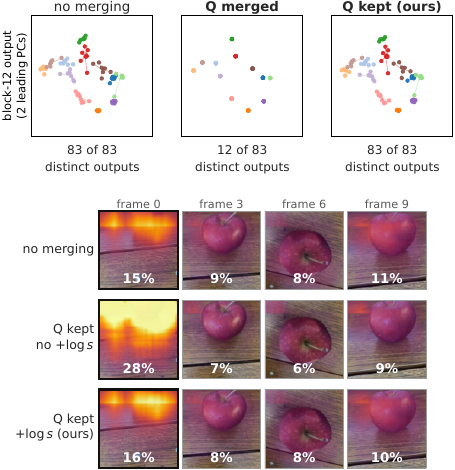}
  \caption{\textbf{Two sources of accuracy degradation in existing token-merging
  methods.} \emph{Top:} Query merging collapses the block-12 outputs of $83$ tokens into
  $12$ shared outputs, whereas retaining all queries while merging only K/V preserves the
  output diversity of the unmerged model. \emph{Bottom:} Omitting the $+\log s$ correction
  disproportionately allocates attention to the first frame. Restoring the correction
  recovers an attention distribution close to that of the unmerged model.}
  \label{fig:query}
\end{figure}

%% file: sec/5_method.tex
\section{GPA-VGGT}
\label{sec:method}

\subsection{Query-Preserving \rev{K/V} Merging with \rev{Token-Size} Compensation}
\label{sec:method:merge}

We merge keys and values while preserving the full query sequence. Each token therefore
attends to the reduced key set using its own query and retains an individual output. In
contrast to FastVGGT~\cite{fastvggt}, which computes attention over
$N_\text{merged} \times N_\text{merged}$ tokens, our formulation has complexity
$N \times N_\text{merged}$. This increases the attention cost by a factor of
$N / N_\text{merged}$ relative to merging queries, but avoids the accuracy loss caused
by replacing token-specific outputs with shared ones. For the remaining configurations,
we follow FastVGGT, using the same grouping rule and merge ratio of $0.9$.

After K/V merging, the remaining approximation is the replacement of $s$ keys by their
mean. \rev{The merged key then has token size $s$ but receives the attention mass of
a single key.} We \rev{correct this bias} by adding $\log s$ to the attention
logit of each merged key, without altering the merged key vector.
\crr{Let $\mathcal{G}_j$ be the set of tokens merged into the $j$-th key,
$s_j=|\mathcal{G}_j|$, and $\bar k_j$, $\bar v_j$ the means of their keys and values.
Every query $q_i$, $i=1,\dots,N$, then produces its own output
\begin{equation}
  o_i=\sum_{j=1}^{N_\text{merged}}
  \mathrm{softmax}_j\!\Big(\frac{q_i^\top \bar k_j}{\sqrt{d}}+\log s_j\Big)\,\bar v_j ,
  \label{eq:kvmerge}
\end{equation}
which equals unmerged attention exactly when the keys within each group are identical.}

\input{fig/3_crossframe}

\input{fig/4_overview}

Accuracy depends primarily on preserving token-specific queries and correctly
normalizing merged keys. Salient-token protection becomes redundant
because all queries are already retained. Removing it simplifies the method
without sacrificing accuracy, while also eliminating its overhead and the
double-counting of destination tokens in the original implementation.

\subsection{Pooled Cross-Frame Broadcast for Early Layers}
\label{sec:method:early}

We retain cross-frame context in the early blocks through a compact \rev{shared
cross-frame K/V set}. Each query attends to three disjoint groups: patch tokens from its own frame, the
camera and four register tokens from all frames, and pooled patch summaries from all
frames. Pooling keeps the shared set compact, while including each frame's own summaries
makes the set identical across frames. It can therefore be constructed once and reused
without masks or per-frame K/V copies.
\crr{For frame $f$, the key set is
$\mathcal{K}_f=[K^\text{patch}_f;\,K^\text{special}_{1:F};\,\bar K^\text{pool}_{1:F}]$, and the
value set is built in the same way.}

We construct content-adaptive pooled summaries using cluster-h. We initialize
representative locations on a stride-$8$ grid, assign each patch token to the
representative with the highest cosine similarity, and average the assigned keys and
values to form each summary. Unlike fixed-window average pooling, this groups patches
with similar content rather than merely nearby locations. Since multi-head attention
splits each $1024$-dimensional token into $16$ heads of dimension $64$, cluster-h
performs this assignment independently for each head using its head-specific features.
Thus, a patch can be grouped differently across heads according to the information each
head captures.
Each pooled summary represents $s$ patch tokens but enters attention as a
single key, so we \rev{apply the same $+\log s$ correction}.

\input{tab/1_pose_main}
\subsection{Core-\rev{Block} Sensitivity-Aware Merging}
\label{sec:method:core}

The aggregator is not uniformly tolerant to merging: blocks $13$--$16$ are substantially
more sensitive than the remaining merged blocks, and errors introduced in this range are
amplified rather than corrected by subsequent blocks. This sensitivity follows from the
attention patterns. In the middle blocks, queries attend sharply to a small number of
keys, particularly cross-view correspondences; averaging keys can therefore dilute the
information carried by the most relevant key. In later blocks, attention is more diffuse
and primarily directed toward global sink tokens, making merging less disruptive. We
consequently use a merge ratio of $0.2$ for blocks $13$--$16$ and retain the ratio of
$0.9$ for all other blocks in the merging window.

\subsection{Gaussian Random Projection for Efficient Matching}
\label{sec:method:matching}

Token matching incurs additional overhead because it determines the source--destination
assignments before merging. Assigning each source token to its most similar destination
requires an $O(\text{src} \times \text{dst} \times C)$ argmax operation, where $C$
denotes the feature dimension. In the accelerated model with the preceding components
applied, matching accounts for approximately $18\%$ of the forward-pass time for a
$200$-frame sequence when using the full $C=1024$ representation.

We reduce matching cost by projecting the matching features into a lower-dimensional
space using a random Gaussian matrix and performing cosine normalization and argmax in
that space\crr{. Each source token $i$ is assigned to the destination
\begin{equation}
  d(i)=\arg\max_{j\in\mathcal{D}}\,\cos\!\big(P^\top x_i,\,P^\top x_j\big),
  \label{eq:rpmatch}
\end{equation}
where $x_i\in\mathbb{R}^{1024}$ is its matching feature, $\mathcal{D}$ the destination set,
and $P\in\mathbb{R}^{1024\times128}$ a random Gaussian matrix shared across all blocks and inputs.} Reducing the feature dimension from $1024$
to $128$ lowers the cost of matching by approximately $7$--$8\times$.

The similarity scores are used only to select destination indices and are discarded
thereafter; merging averages the tokens assigned to each group. Even when
projection changes the selected destination, it typically selects an alternative among
near-tied candidates: token features are highly redundant, and the most similar
destination is rarely unique. Thus, random projection may reorder nearly equivalent
matches without producing a fundamentally different assignment. Since merging requires
each token to be grouped with a sufficiently similar token rather than one specific
destination, the projection reduces matching cost while preserving similarity-based
token groups.

\subsection{Custom Kernels}
\label{sec:method:impl}

\crr{Two custom Triton~\cite{triton} kernels implement the design: a
FlashAttention~\cite{flashattention}-style kernel that adds the $+\log s$ bias inside its
online-softmax loop, since passing it as a mask makes PyTorch~\cite{pytorch} SDPA fall back
to a slower backend, and a matching kernel that computes the source--destination argmax
without storing the score matrix. Both are numerically equivalent to their PyTorch
references; Appendix~\ref{sec:supp:kernels} gives details.}
\input{fig/5_broccoli}

%% file: fig/3_crossframe.tex
\begin{figure}[!t]
  \centering
  \includegraphics[width=\linewidth]{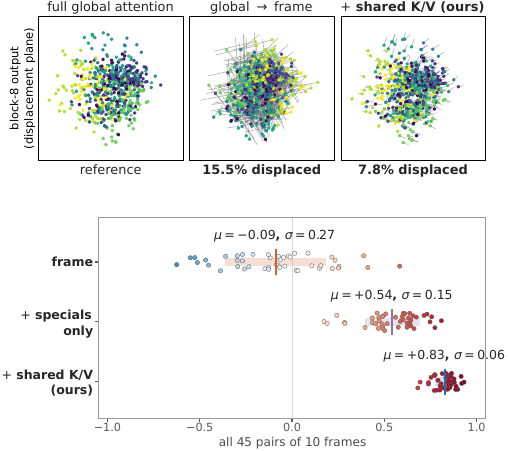}
  \caption{\textbf{Effect of cross-frame context in the early blocks.}
  \emph{Top:} Replacing global attention with \rev{frame} attention substantially
  displaces the block-8 output field from the full-global-attention reference, whereas
  \crr{the} \rev{shared cross-frame K/V set} largely preserves it. \emph{Bottom:} Pairwise alignment of \rev{per-frame} drift
  directions. Cross-frame special tokens partially align the
  drift, and image-token summaries further reduce its variation, preserving a coherent
  geometric reference across views.}
  \label{fig:crossframe}
\end{figure}

%% file: fig/4_overview.tex
\begin{figure*}[t]
  \centering
  \includegraphics[width=\linewidth]{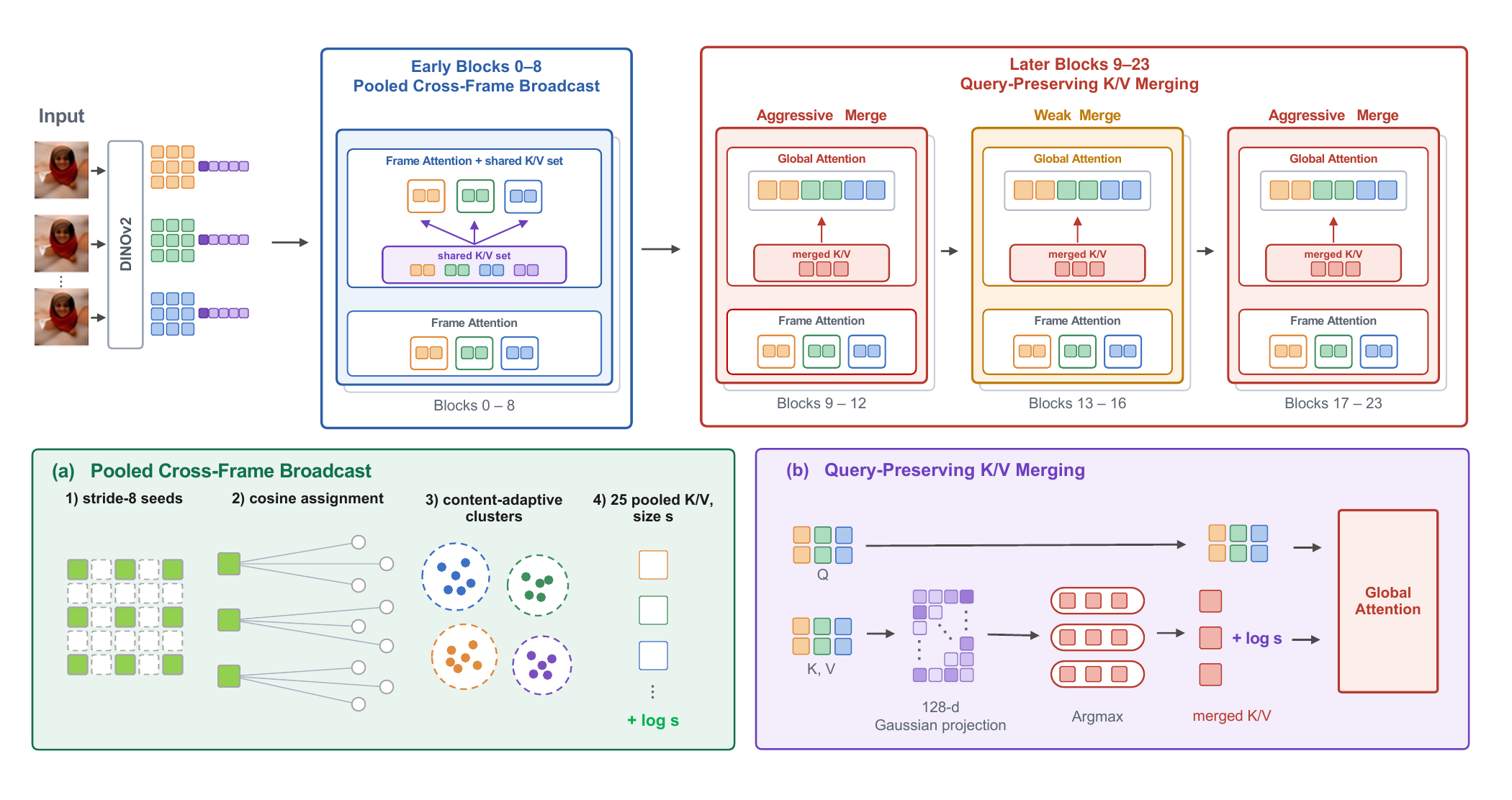}
  \caption{\textbf{Overview of GPA-VGGT.} Early blocks $0$--$8$ replace global
  attention with frame attention augmented by a shared cross-frame K/V \rev{set} containing
  special tokens and content-adaptive pooled image-token summaries. \rev{Later} blocks
  $9$--$23$ preserve all queries and merge only K/V tokens, using a weaker merge ratio in
  the \rev{core} blocks $13$--$16$. \textbf{(a)}~Pooled summaries are constructed by
  assigning patch tokens to stride-8 seeds according to cosine similarity and averaging
  the K/V tokens within each content-adaptive cluster; the summary size determines the
  \(+\log s\) correction. \textbf{(b)}~K/V matching is computed after a
  $128$-dimensional Gaussian projection, while all queries are retained for global
  attention.}
  \label{fig:overview}
\end{figure*}

%% file: tab/1_pose_main.tex
\begin{table*}[!t]
  \centering\small\setlength{\tabcolsep}{0pt}
  \begin{tabular*}{\linewidth}{@{\extracolsep{\fill}}lcccccccc@{}}
    \toprule
    & \multicolumn{4}{c}{CO3D} & \multicolumn{4}{c@{}}{\rev{RE10K}} \\
    \cmidrule(lr){2-5}\cmidrule(l){6-9}
    & AUC@3 & AUC@5 & AUC@10 & AUC@30 & AUC@3 & AUC@5 & AUC@10 & AUC@30 \\
    \midrule
    \input{tab/gen/pose_main}\\
    \bottomrule
  \end{tabular*}
  \caption{\textbf{Camera-pose accuracy on CO3D and \rev{RE10K}}, with ten frames
  per sequence and relative pose errors computed over all frame pairs.
  Bold and underline mark the best and second-best value in each column,
  excluding vanilla VGGT.}
  \label{tab:pose_main}
\end{table*}

%% file: tab/gen/pose_main.tex
    vanilla VGGT               & 57.49 & 68.88 & 80.66 & 91.71 & 44.78 & 56.67 & 70.79 & 85.76 \\
    FastVGGT                   & 41.56 & 55.97 & 72.61 & 88.38 & 37.56 & 50.52 & 66.38 & 83.69 \\
    FasterVGGT                 & 16.05 & 33.80 & 60.00 & 83.91 & 22.00 & 34.09 & 51.47 & 73.85 \\
    AVGGT(2)                   & \underline{51.87} & \underline{64.55} & \underline{78.02} & \underline{90.65} & \textbf{44.68} & \textbf{56.49} & \textbf{70.66} & \textbf{85.70} \\
    AVGGT(4)                   & 47.08 & 60.75 & 75.72 & 89.72 & 42.53 & 54.78 & 69.50 & 85.21 \\
    \textbf{GPA-VGGT}          & \textbf{56.94} & \textbf{68.54} & \textbf{80.41} & \textbf{91.57} & \underline{44.45} & \underline{56.33} & \underline{70.46} & \underline{85.54}%

%% file: fig/5_broccoli.tex
\begin{figure*}[t]
  \centering
  \includegraphics[width=\linewidth]{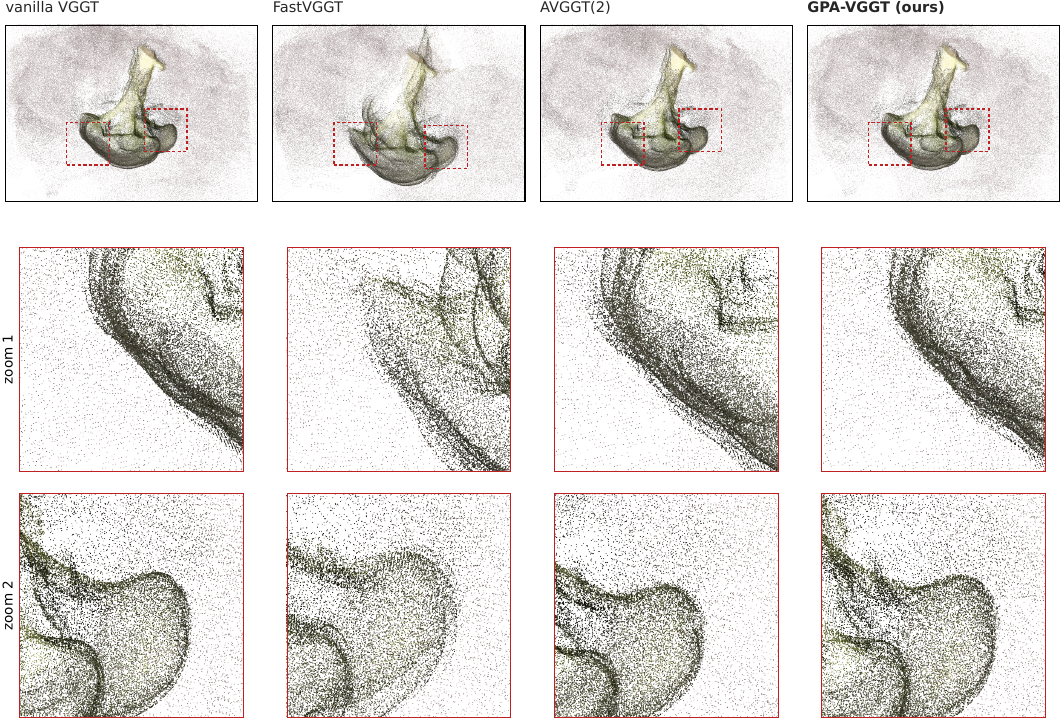}
  \caption{\textbf{Qualitative reconstruction under estimated camera poses.} Each column
  shows the point cloud produced by the corresponding method, and red boxes indicate the
  enlarged regions below. GPA-VGGT most closely matches the geometry reconstructed by
  vanilla VGGT.}
  \label{fig:qual}
\end{figure*}

%% file: sec/6_experiments.tex
\section{Experiments}
\label{sec:exp}

\rev{We organize our evaluation around four questions:
(1) Does GPA-VGGT preserve camera-pose accuracy under strict thresholds, and does it
keep doing so as the sequence grows longer?
(2) Does the same preservation carry over to other tasks?
(3) How does its runtime speedup compare with existing acceleration
methods?
(4) Which components are responsible for accuracy recovery and acceleration?}

\subsection{Experimental \crr{Setup}}
\label{sec:exp:setup}
\input{tab/2_longseq}

We follow the $\pi^3$~\cite{pi3} evaluation protocol for both tasks. For camera pose
estimation, we evaluate on CO3D~\cite{co3d} and \rev{RealEstate10K (RE10K)}~\cite{re10k} with ten frames per
sequence, compute relative pose errors over all frame pairs, and report AUC at the
standard thresholds as well as the stricter AUC@3. We additionally evaluate
long-sequence behavior on CO3D using \rev{$64$, $100$, and $200$} input frames.

For point-map reconstruction, we evaluate on 7-Scenes~\cite{sevenscenes} and
NRGBD~\cite{nrgbd} using accuracy, completeness, and normal consistency, computed over
pixels valid under the ground-truth mask. We follow the sparse and dense
\rev{frame samplings of} $\pi^3$ and report the two settings separately.
The sparse \rev{samplings} contain $26$ frames across $9$ NRGBD sequences and $86$ frames
across $18$ 7-Scenes sequences, corresponding to $1$--$5$ selected frames per sequence;
the dense \rev{samplings} contain $114$ and $426$ frames, respectively, corresponding to
$4$--$25$ frames per sequence.

We retain the original VGGT~\cite{vggt} architecture with $L=24$ alternating frame-attention and
global-attention layers. All methods use the same VGGT released
checkpoint without training or fine-tuning. Experiments use \texttt{bfloat16} autocast
and are conducted on a single NVIDIA RTX A6000 GPU with $48$\,GiB of memory, using
PyTorch 2.11 and CUDA 12.8.

\crr{As the official} AVGGT~\cite{avggt} \crr{code is unreleased}, we
reimplemented the method from its specification. \rev{\crr{It} reproduces the
reported qualitative trend on RE10K.} \crr{Absolute AUC values are not comparable,
since \crr{its RE10K subset and sampling are not public}.} 

\subsection{Camera \crr{Pose}}
\label{sec:exp:pose}

Table~\ref{tab:pose_main} highlights that under AUC@30, most
accelerated methods remain close to vanilla VGGT; under the stricter AUC@3 metric,
however, their performance gaps become more pronounced.

The results also differ across the two benchmarks. AVGGT(2) \crr{stays close to} vanilla VGGT on \rev{RE10K} but incurs a much larger AUC@3 drop on CO3D. This difference reflects AVGGT's fixed-stride K/V subsampling, in contrast to GPA-VGGT's content-based K/V merging. \rev{RE10K} consists of temporally adjacent indoor views with high overlap and predominantly smooth surfaces, making positional subsampling effective. In contrast, CO3D contains wider-baseline views of detailed objects, where content-aware selection is needed to preserve the relevant K/V information. Accordingly, GPA-VGGT preserves accuracy across both benchmarks.

Table~\ref{tab:longseq} shows that this trend holds across sequence lengths. \rev{GPA-VGGT's} AUC@3 gap from vanilla VGGT does not widen as the sequence length increases, while competing methods remain less accurate.

\input{tab/3_mvrecon}
Figure~\ref{fig:qual} qualitatively compares the reconstructed
geometry on the same scene.
FastVGGT shows the object boundary as two separate edges and makes the nearby surface blurry, and AVGGT(2) reconstructs the same region as two nearby surface layers. At the scale of the full object it looks similar to vanilla VGGT, and only the zoomed-in view reveals a thinner surface behind those layers. GPA-VGGT preserves one clear boundary and remains visually close to vanilla VGGT. \rev{FastVGGT remains close to vanilla VGGT under AUC@30 but degrades
substantially under AUC@3. The qualitative reconstructions are consistent with this
gap, indicating that AUC@3 more directly reflects visible geometric degradation.}

\subsection{\crr{Point-Map Reconstruction}}
\label{sec:exp:mvrecon}

Table~\ref{tab:mvrecon} shows that GPA-VGGT preserves point-map reconstruction quality
across datasets and sampling densities. It is closest to vanilla VGGT on
dense NRGBD and sparse 7-Scenes, while maintaining near-vanilla normal consistency in
every setting. AVGGT(2) achieves the best results on sparse NRGBD, but its fixed-stride
subsampling degrades more substantially on dense NRGBD and sparse 7-Scenes. FastVGGT and
FasterVGGT also show consistent degradation, particularly in normal consistency.

\subsection{\rev{Runtime}}
\label{sec:exp:latency}

Table~\ref{tab:latency} reports end-to-end runtime across sequence lengths. At $10$ frames, no method
exceeds a $1.2\times$ speedup over vanilla VGGT, whereas all methods exceed $2.3\times$
at $200$ frames. At $500$ frames, FasterVGGT \crr{runs out of memory}, while the remaining methods achieve speedups between $2.8\times$ and
$4.6\times$. AVGGT(4) is the fastest method beyond $32$ frames, reaching $4.60\times$ at
$500$ frames, compared with $3.24\times$ for GPA-VGGT; this additional speedup, however,
accompanies a substantially larger accuracy loss.

\raggedbottom

\input{tab/4_latency}

\subsection{Ablation}
\label{sec:exp:ablation}

Table~\ref{tab:ablation} shows a clear division of roles among the design components.
The second rung is the early global-to-frame conversion AVGGT~\cite{avggt} proposes,
applied on its own: it provides the initial runtime reduction, cutting $27.7$ of the
$86.8$ seconds, but lowers \crr{CO3D AUC@3 by $3.06$ points}.
Pooled cross-frame broadcast then restores the missing cross-view context and recovers
\rev{$2.82$} of those points for $0.7$ additional seconds. Query-preserving K/V merging then provides the main
additional acceleration without substantially compromising accuracy. Random-projection
matching and sensitivity-aware merging further reduce computation overhead with only a
small additional effect on accuracy.
\crr{The same design transfers to $\pi^3$~\cite{pi3} and VGGT-$\Omega$~\cite{vggtomega},
reaching $2.36\times$ and $1.84\times$ speedups at $200$ frames while largely preserving
the AUC@3 of their vanilla models on CO3D and RE10K
(Appendix~\ref{sec:supp:transfer}).}

\input{tab/5_ablation}

%% file: tab/2_longseq.tex
\begin{table}[!t]
  \centering\footnotesize\setlength{\tabcolsep}{3.5pt}
  \begin{tabular}{@{}lccc@{\hskip 10pt}ccc@{}}
    \toprule
    & \multicolumn{3}{c}{AUC@3 $\uparrow$} & \multicolumn{3}{c}{AUC@30 $\uparrow$} \\
    \cmidrule(r){2-4}\cmidrule(l){5-7}
    & \rev{64} & \rev{100} & \rev{200} & \rev{64} & \rev{100} & \rev{200} \\
    \midrule
    \input{tab/gen/longseq}\\
    \bottomrule
  \end{tabular}
  \caption{\textbf{Camera-pose accuracy on CO3D at \crr{$64$, $100$, and $200$} input
  frames.} Bold and underline mark the best and second-best value in each column,
  excluding vanilla VGGT.}
  \label{tab:longseq}
\end{table}

%% file: tab/gen/longseq.tex
    vanilla VGGT         & 62.57 & 62.45 & 61.47 & 92.58 & 92.59 & 92.42 \\
    FastVGGT             & 55.18 & 55.97 & 56.28 & 91.33 & 91.54 & 91.68 \\
    FasterVGGT           & 20.66 & 20.40 & 19.59 & 85.62 & 85.57 & 85.42 \\
    AVGGT(2)             & \underline{58.49} & \underline{58.59} & \underline{57.71} & \underline{91.94} & \underline{92.03} & \underline{91.85} \\
    AVGGT(4)             & 54.27 & 54.76 & 54.40 & 91.35 & 91.50 & 91.45 \\
    \textbf{GPA-VGGT}    & \textbf{62.61} & \textbf{62.55} & \textbf{61.43} & \textbf{92.57} & \textbf{92.59} & \textbf{92.41}%

%% file: tab/3_mvrecon.tex
\begin{table*}[!t]
  \centering\footnotesize
  \setlength{\tabcolsep}{0pt}
  \begin{tabular*}{\linewidth}{@{\extracolsep{\fill}}lcccccccccccc@{}}
    \toprule
    & \multicolumn{3}{c}{NRGBD sparse} & \multicolumn{3}{c}{NRGBD dense}
    & \multicolumn{3}{c}{7-Scenes sparse} & \multicolumn{3}{c@{}}{7-Scenes dense} \\
    \cmidrule(r){2-4}\cmidrule(lr){5-7}\cmidrule(lr){8-10}\cmidrule(l){11-13}
    & Acc $\downarrow$ & Comp $\downarrow$ & NC $\uparrow$
    & Acc $\downarrow$ & Comp $\downarrow$ & NC $\uparrow$
    & Acc $\downarrow$ & Comp $\downarrow$ & NC $\uparrow$
    & Acc $\downarrow$ & Comp $\downarrow$ & NC $\uparrow$ \\
    \midrule
    \input{tab/gen/mvrecon}\\
    \bottomrule
  \end{tabular*}
  \caption{\textbf{Point-map reconstruction on NRGBD and 7-Scenes}, under the sparse
  and dense \rev{frame samplings of} $\pi^3$.
  Bold and underline mark the best and second-best value in each column,
  excluding vanilla VGGT.}
  \label{tab:mvrecon}
\end{table*}

%% file: tab/gen/mvrecon.tex
    vanilla VGGT               & 0.0510 & 0.0665 & 0.8843 & 0.0144 & 0.0152 & 0.8743 & 0.0432 & 0.0557 & 0.7337 & 0.0215 & 0.0259 & 0.6673 \\
    FastVGGT                   & 0.0560 & 0.0819 & 0.8601 & 0.0308 & 0.0328 & 0.8302 & \underline{0.0510} & \underline{0.0696} & \underline{0.7215} & \textbf{0.0195} & \textbf{0.0235} & 0.6646 \\
    FasterVGGT                 & 0.1551 & 0.2173 & 0.7749 & 0.0759 & 0.0642 & 0.7988 & 0.0994 & 0.1527 & 0.6637 & 0.0246 & 0.0254 & \textbf{0.6695} \\
    AVGGT(2)                   & \textbf{0.0355} & \textbf{0.0396} & \textbf{0.8981} & \underline{0.0154} & \underline{0.0157} & \underline{0.8562} & 0.0613 & 0.0765 & 0.7115 & 0.0210 & 0.0257 & 0.6653 \\
    AVGGT(4)                   & 0.0465 & 0.0554 & 0.8818 & 0.0207 & 0.0182 & 0.8466 & 0.0656 & 0.0866 & 0.7063 & 0.0218 & \underline{0.0248} & 0.6665 \\
    \textbf{GPA-VGGT}          & \underline{0.0414} & \underline{0.0476} & \underline{0.8941} & \textbf{0.0143} & \textbf{0.0151} & \textbf{0.8714} & \textbf{0.0474} & \textbf{0.0600} & \textbf{0.7309} & \underline{0.0204} & 0.0253 & \underline{0.6674}%

%% file: tab/4_latency.tex
\begin{table}[!t]
  \centering\footnotesize
  \setlength{\tabcolsep}{2pt}
  \renewcommand{\arraystretch}{1.10}
  \begin{tabular*}{\linewidth}{@{\extracolsep{\fill}}lrrrrrr@{}}
    \toprule
    & \multicolumn{6}{c@{}}{number of input frames} \\
    \cmidrule(l){2-7}
    time (s) $\downarrow$ & 10 & 32 & 64 & 100 & 200 & 500 \\
    \midrule
    \input{tab/gen/latency}\\
    \bottomrule
  \end{tabular*}
  \caption{\textbf{End-to-end \rev{runtime}} for the whole forward pass, camera and
  depth heads included, on a single A6000.}
  \label{tab:latency}
\end{table}

%% file: tab/gen/latency.tex
    vanilla VGGT               & 0.84 & 3.84 & 12.46 & 24.01 & 86.78 & 513.78 \\
    FastVGGT                   & 0.85 & 2.81 & 6.46 & 11.71 & 33.91 & 169.43 \\
    FasterVGGT                 & 0.78 & 2.66 & 6.18 & 11.29 & 32.27 & OOM \\
    AVGGT(2)                   & 0.78 & 2.76 & 6.70 & 12.66 & 37.02 & 184.03 \\
    AVGGT(4)                   & 0.76 & 2.51 & 5.58 & 9.86 & 25.82 & 111.74 \\
    \textbf{GPA-VGGT}          & 0.78 & 2.76 & 6.49 & 11.92 & 33.65 & 158.63%

%% file: tab/5_ablation.tex
\begin{table}[!t]
  \centering\footnotesize
  \setlength{\tabcolsep}{2.5pt}
  \begin{tabular*}{\linewidth}{@{\extracolsep{\fill}}lrcc@{}}
    \toprule
                                                & time $\downarrow$ & \multicolumn{2}{c@{}}{AUC@3 $\uparrow$} \\
                                                \cmidrule(lr){3-4}
                                                & $200$f & CO3D & RE10K \\
    \midrule
    \input{tab/gen/ablation}\\
    \bottomrule
  \end{tabular*}
  \caption{\textbf{\rev{Ablation of GPA-VGGT components.}} Each row adds one
  component to the row above it.}
  \label{tab:ablation}
\end{table}

%% file: tab/gen/ablation.tex
    vanilla VGGT                                 & 86.8\,s & 57.49 & 44.78 \\
    $+$ \rev{frame attention in early blocks}    & 59.1\,s & 54.43 & 44.45 \\
    $+$ pooled cross-frame broadcast             & 59.8\,s & 57.25 & 45.20 \\
    $+$ query-preserving K/V merging             & 40.1\,s & 57.13 & 44.45 \\
    \rev{\quad on blocks 9--12 and 17--23}       &         &       &       \\
    $+$ random-projection matching               & 35.5\,s & 57.12 & 44.47 \\
    $+$ sensitivity-aware core, 13--16           & 33.6\,s & 56.94 & 44.45%

%% file: sec/8_conclusion.tex
\section{Conclusion}
\label{sec:conclusion}

\rev{We identify three sources of accuracy degradation in accelerated VGGT:
information loss from query merging, incorrect attention mass after K/V
merging, and the removal of early cross-frame context. GPA-VGGT addresses these
issues through \crr{Query-Preserving K/V Merging with $+\log s$ correction, Pooled
Cross-Frame Broadcast, and low-dimensional matching}.
Across diverse multi-view geometry evaluations, GPA-VGGT reduces
inference time while preserving \crr{vanilla VGGT's accuracy under both standard
and strict metrics}.}

%% file: supp/0_supp.tex
\clearpage
\maketitlesupplementary
\appendix

\renewcommand{\thefigure}{S\arabic{figure}}
\renewcommand{\thetable}{S\arabic{table}}
\flushbottom
\setcounter{figure}{0}
\setcounter{table}{0}

\section{Custom Triton Kernels}
\label{sec:supp:kernels}

\subsection{Attention with a Per-Key Additive Bias}
\label{sec:supp:kernels:bias}

The standard attention interface treats the $+\log s$ correction as a mask and therefore
falls back to a slower implementation. For a $200$-frame sequence, this fallback takes
$2{,}142$\,ms, compared with $948$\,ms for unbiased fused attention---more time than key
merging saves in the merged global-attention block.

Because the correction depends only on the key index, our Triton kernel adds it directly
to each score tile inside the online-softmax loop, preserving the streaming structure of
FlashAttention~\cite{flashattention}. We avoid masked loads by clamping out-of-range key
indices and assigning duplicate columns a $-10^{9}$ bias. This avoids padding and copying
K/V tensors; only the small bias vector is padded. The bias is also pre-scaled on the
host, avoiding conversion within the inner loop. For a $200$-frame sequence, our
kernel takes $984$\,ms: only $4\%$ slower than unbiased fused attention and
$2.18\times$ faster than the fallback.

\subsection{Memory-Efficient K/V Matching}
\label{sec:supp:kernels:sim}

Bipartite matching requires only the row-wise maximum similarity and argmax index, not
the full similarity matrix. Materializing the full matrix would require $27.25$\,GiB for a
$200$-frame sequence. Even a chunked PyTorch~\cite{pytorch} reference, which materializes
$5{,}000$ rows at a time, requires a $1.20$\,GiB temporary buffer. Our Triton kernel keeps
each score tile in registers and writes only the two required outputs, reducing this
temporary memory to $0.003$\,GiB.

The kernel trades a modest increase in \rev{runtime} for substantially lower memory
use: \rev{the similarity kernel itself} takes $482$\,ms against $330$\,ms for the
chunked reference \rev{at this shape}. This overhead of roughly $150$\,ms
is small relative to the memory saved, since materializing similarity tiles
incurs a large transient VRAM cost and can become the limiting factor for long sequences.

The PyTorch reference delegates the $1{,}024$-dimensional dot-product reduction to its
matrix-multiplication backend. In our fused kernel, which additionally computes the
row-wise maximum and argmax, we partition the feature dimension into smaller tiles and
accumulate partial dot products.

\section{Peak VRAM Footprint}
\label{sec:supp:vram}

\input{tab/s2_memory}

Table~\ref{tab:supp:memory} reports the peak VRAM of the full forward pass, using the
VRAM-efficient VGGT implementation introduced by FastVGGT~\cite{fastvggt} as the vanilla
baseline. GPA-VGGT remains within $0.02$\,GiB of vanilla VGGT across all sequence
lengths and AVGGT matches it exactly, whereas at $500$ frames FastVGGT uses
$6.02$\,GiB more VRAM; FasterVGGT already uses $13.72$\,GiB more at $400$ frames and
exceeds device memory at $500$.

Although K/V merging reduces the K/V sequence length in global attention, the
full-model peak is set by the unchanged feed-forward sublayer within a global-aggregator
block. Moreover, FlashAttention~\cite{flashattention} does not materialize an
$N \times N$ score matrix, so the memory used by the modified global-attention sublayer
is already linear in the number of tokens. GPA-VGGT therefore reduces attention
computation without changing peak VRAM, while keeping attention memory nearly identical
to vanilla VGGT and below the feed-forward-memory peak. In contrast, FastVGGT merges
queries as well as K/V tokens and subsequently restores the original token resolution.
The associated full-length token copies and unmerging buffers raise its attention-memory
peak above that of the feed-forward sublayer.

AVGGT incurs essentially no additional peak-memory overhead. Its subsampled global attention drops keys rather than
copying tokens back, so it allocates neither FastVGGT's unmerging buffers nor any
workspace that scales with the dropped set, and its peak equals vanilla VGGT's at every
sequence length and at both subsampling settings.

\section{Pooling for the Shared Cross-Frame K/V Set}
\label{sec:supp:pooling}

Each frame's queries attend to local K/V tokens, shared special tokens, and pooled
image-token summaries. The first two groups are held fixed in this ablation; this section studies
how to construct the summaries. Table~\ref{tab:supp:pooling} compares pooling strategies
on CO3D using a fixed $8 \times 8$ pooling window. Fixed-window mean pooling performs worst,
while content-based clustering is more effective. Per-head clustering performs best and
is used in our model.

\input{tab/s3_pooling}
\input{tab/s4_mergeop}

\section{Merge-Operator Ablation}
\label{sec:supp:bcastq}

The merge operator independently determines whether to merge queries with K/V tokens and
whether to correct each merged key's logit by its \rev{token size}.
Table~\ref{tab:supp:mergeop} isolates these choices under an otherwise fixed
configuration. Merging queries forces all tokens within a group to share the same output,
whereas preserving queries retains within-group diversity but does not address the
\rev{attention-mass bias} caused by K/V merging. The $+\log s$ correction \rev{restores
the mass of the tokens each merged key represents} and provides the larger accuracy gain. Together, query
preservation and \rev{the $+\log s$ correction} nearly recover no-merging accuracy, with the
benefit becoming larger at higher merge ratios.

\section{Selection of the Core Blocks}
\label{sec:supp:core}

\input{fig/s1_core_selection}

We choose the core blocks by measuring how much merging changes the
aggregator's final representation. For each candidate set, we leave those blocks unmerged and apply merging
to every other block. We measure the relative Frobenius-norm deviation between the final
global-block outputs of the merged and unmerged models on the special tokens, and
normalize it by the deviation obtained when all blocks are merged. Because this
measurement uses no ground-truth annotations, the core blocks are selected without
reference to task labels.

We evaluate block sets rather than individual blocks. Simply ranking individual
blocks and selecting the top $k$ does not reliably find the best set. A
single-block measurement considers one block while all other blocks remain
unmerged, whereas the full model applies merging across multiple blocks. Merging
in one block can therefore change the token representations and
source--destination assignments used by later blocks.

For VGGT, this criterion selects blocks $13$--$16$. Figure~\ref{fig:supp:core}
shows that this set gives the smallest output change on all four datasets. No
three-block set matches this result. We apply the same procedure to $\pi^3$ and VGGT-$\Omega$;
Figure~\ref{fig:supp:coretransfer} shows the resulting core-block selections,
which are used in Section~\ref{sec:supp:transfer}.

\section{Matching Feature Dimension}
\label{sec:supp:matchdim}

Table~\ref{tab:supp:matchdim} evaluates the fixed random projection used for K/V
matching. Projecting the matching features from $1{,}024$ to $128$ dimensions lowers the
cost of each matching call from $469$\,ms to $64$\,ms. Although only $76.6\%$ of source
tokens select the same destination as under the full-dimensional metric, the selected
destinations lose only $0.56\%$ similarity relative to the full-metric optimum,
indicating that most disagreements occur among near-equivalent candidates. \rev{We
therefore use $128$ dimensions, the smallest projection size that maintains
accuracy close to the full-dimensional metric.}

\input{tab/s6_matchdim}

\section{Comparison with AVGGT}
\label{sec:supp:avggt}

GPA-VGGT and AVGGT~\cite{avggt} share a two-stage structure: both replace early
global-attention blocks with frame attention and reduce the K/V set in later blocks
while preserving all queries. In the early stage, AVGGT restricts attention to
within-frame tokens, whereas GPA-VGGT augments each frame with a pooled shared
cross-frame K/V set. In the later stage, AVGGT retains K/V tokens at fixed spatial
strides, whereas GPA-VGGT selects them through bipartite merging. At an equal token
budget, \rev{AVGGT summarizes all removed tokens with a single unweighted mean K/V
pair, whereas GPA-VGGT aggregates each removed token into the retained token it most
closely matches and restores the corresponding attention mass} through
the $+\log s$ correction. The $128$-dimensional Gaussian projection reduces matching overhead,
enabling GPA-VGGT to retain fewer K/V tokens than AVGGT(2) while achieving both lower
runtime and higher accuracy.

\input{tab/s7_transfer}
\input{fig/s2_core_transfer}

\section{Transfer to Other Backbones}
\label{sec:supp:transfer}

The proposed design applies to backbones that alternate frame attention and
global attention over a token layout comprising special tokens followed by a patch grid. Both
$\pi^3$~\cite{pi3} and VGGT-$\Omega$~\cite{vggtomega} follow this structure. We adjust only model-specific properties,
including the block layout, the number of special tokens per frame, and patch size. Each
adaptation uses the same $128$-dimensional Gaussian projection for K/V matching.

Table~\ref{tab:supp:transfer:pi3} reports the incremental ablation on $\pi^3$. We
retain the early-block setting used by AVGGT: global attention in
blocks $0$--$9$ is replaced by frame attention, and K/V merging is applied in blocks
$10$--$17$, with a lower merge ratio in core blocks $13$--$15$. However, $\pi^3$
requires a lower overall merge ratio than VGGT because it has fewer global-attention
blocks. Each merged block therefore has a larger effect on the information propagated
through the remaining global-attention blocks.

Table~\ref{tab:supp:transfer:omega} reports the same ablation on VGGT-$\Omega$, which
contains five register blocks at raw indices $2$, $6$, $9$, $14$, and $20$ that do not
perform cross-frame attention over patch tokens. We therefore replace global
attention with frame attention in raw blocks $0$, $1$, $3$, $4$, $5$, $7$, and $8$, and
apply K/V merging in the subsequent non-register blocks, with a lower merge ratio in raw
blocks $12$, $13$, $15$, and $16$. As in $\pi^3$, patch tokens
participate in global attention in fewer blocks than in VGGT, requiring a lower overall
merge ratio.

On both backbones, the shared cross-frame \rev{K/V} set mitigates the accuracy loss
caused by replacing early global attention with frame attention. The adopted configurations accelerate all three
backbones while retaining most of the accuracy of their vanilla models.

\input{supp/I_qualitative}

%% file: tab/s2_memory.tex
\begin{table}[t]
  \centering\scriptsize
  \setlength{\tabcolsep}{3pt}
  \renewcommand{\arraystretch}{1.05}
  \begin{tabular}{@{}rrrrrr@{}}
    \toprule
    frames & vanilla & FastVGGT & FasterVGGT & AVGGT & GPA-VGGT \\
    \midrule
     10 &  4.58 &  4.58 &  4.58 &  4.58 &  4.58 \\
     32 &  5.12 &  5.12 &  5.12 &  5.12 &  5.12 \\
     64 &  6.08 &  6.85 &  7.48 &  6.08 &  6.08 \\
    100 &  8.27 &  9.48 & 10.50 &  8.27 &  8.28 \\
    \addlinespace[3pt]
    200 & 14.37 & 16.78 & 19.46 & 14.37 & 14.38 \\
    300 & 20.47 & 24.08 & 29.43 & 20.47 & 20.48 \\
    400 & 26.57 & 31.38 & 40.29 & 26.57 & 26.58 \\
    500 & 32.67 & 38.69 & OOM & 32.68 & 32.69 \\
    \bottomrule
  \end{tabular}
  \caption{\textbf{Peak GPU memory} (GiB) during the full forward pass on a single
  A6000, measured using \texttt{torch.cuda.max\_memory\_allocated()} after warm-up. Vanilla denotes
  the VRAM-efficient VGGT implementation introduced by FastVGGT, and the AVGGT column
  covers both $\sigma=2$ and $\sigma=4$, which produced identical measurements at every length.}
  \label{tab:supp:memory}
\end{table}

%% file: tab/s3_pooling.tex
\begin{table}[t]
  \centering\footnotesize
  \setlength{\tabcolsep}{0pt}
  \renewcommand{\arraystretch}{1.1}
  \begin{tabular*}{\linewidth}{@{\extracolsep{\fill}}lrrrrr@{}}
    \toprule
    & \new{vanilla} & mean & stride & cluster & \textbf{cluster-h} \\
    \midrule
    CO3D AUC@3 & \rev{$57.49$} & \rev{$56.44$} & \rev{$56.61$} & \rev{$56.90$} & \rev{$56.94$} \\
    $\Delta$ vs.\ \new{vanilla} & -- & \rev{$-1.04$} & \rev{$-0.88$} & \rev{$-0.59$} & \rev{$-0.55$} \\
    \bottomrule
  \end{tabular*}
  \caption{\textbf{Pooling strategy for the shared cross-frame summaries} on CO3D,
  using the same stride-$8$ seed grid and summary-token budget. Bold marks the adopted setting.}
  \label{tab:supp:pooling}
\end{table}

%% file: tab/s4_mergeop.tex
\begin{table}[t]
  \centering\footnotesize
  \setlength{\tabcolsep}{0pt}
  \renewcommand{\arraystretch}{1.05}
  \begin{tabular*}{\linewidth}{@{\extracolsep{\fill}}lccrr@{}}
    \toprule
    & preserve & $+\log s$ & output & CO3D \\
    & Q        & bias      & error  & AUC@3 \\
    \midrule
    no merging            & --         & --         & --    & 57.25 \\
    \multicolumn{5}{@{}l@{}}{\new{\quad (Pooled Cross-Frame Broadcast only)}} \\
    \midrule
    neither component     & --         & --         & 0.394 & 49.93 \\
    query preservation only & \checkmark & --       & 0.374 & 51.09 \\
    $+\log s$ correction only & --      & \checkmark & 0.112 & 55.36 \\
    \textbf{both (adopted)} & \checkmark & \checkmark & \textbf{0.055} & \textbf{56.94} \\
    \bottomrule
  \end{tabular*}
  \caption{\textbf{Merge-operator components} on CO3D, under a fixed merge window
  and merge ratio. Output error is the relative deviation of the merged attention
  output from unmerged attention on the same input. Bold marks the adopted operator.}
  \label{tab:supp:mergeop}
\end{table}

%% file: fig/s1_core_selection.tex
\begin{figure}[t]
  \centering
  \includegraphics[width=\linewidth]{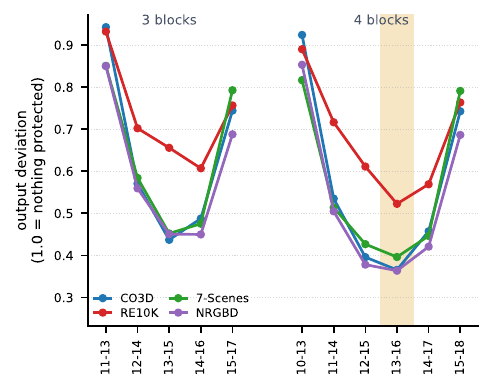}
  \caption{\textbf{Core-block selection on VGGT.} Each point shows how much the
  final global-block output changes from the unmerged model when the indicated
  contiguous block set is protected from merging; lower values mean smaller
  changes. The criterion is computed without ground-truth annotations across
  four datasets. Blocks $13$--$16$ consistently give the smallest change and are
  selected as the core-block set.}
  \label{fig:supp:core}
\end{figure}

%% file: tab/s6_matchdim.tex
\begin{table}[t]
  \centering\footnotesize
  \setlength{\tabcolsep}{3pt}
  \renewcommand{\arraystretch}{1.05}
  \begin{tabular*}{\columnwidth}{@{\extracolsep{\fill}}lrrrrrr@{}}
    \toprule
    dimension & 32 & 64 & \textbf{128} & 256 & 512 & 1024 \\
    \midrule
    matching (ms) & 37 & 38 & \textbf{64} & 106 & 194 & 469 \\
    same destination (\%) & 54.4 & 67.1 & \textbf{76.6} & 82.8 & 87.9 & 100 \\
    similarity loss (\%) & 3.30 & 1.28 & \textbf{0.56} & 0.28 & 0.13 & 0 \\
    \midrule
    CO3D AUC@3 & 56.47 & 56.48 & \textbf{56.94} & 56.87 & 56.89 & 57.09 \\
    \bottomrule
  \end{tabular*}
  \caption{\textbf{Projected feature dimension for K/V matching}, with matching cost
  measured at $200$ input frames and AUC@3 on CO3D. \emph{Same destination} and \emph{similarity loss} are both measured against the full $1{,}024$-dimensional metric. Bold marks the
  adopted dimension.}
  \label{tab:supp:matchdim}
\end{table}

%% file: tab/s7_transfer.tex
\begin{table*}[!t]
  \centering\footnotesize
  \setlength{\tabcolsep}{4pt}
  \renewcommand{\arraystretch}{1.05}
  \begin{tabular}{@{}lrrrrcrrrrcrr@{}}
    \toprule
      & \multicolumn{4}{c}{CO3D AUC@$\tau$} && \multicolumn{4}{c}{RE10K AUC@$\tau$}
      && \multicolumn{2}{c}{$200$ frames} \\
    \cmidrule(lr){2-5}\cmidrule(lr){7-10}\cmidrule(lr){12-13}
      & $3$ & $5$ & $10$ & $30$ && $3$ & $5$ & $10$ & $30$
      && time (s) & speedup \\
    \midrule
    vanilla                        & $58.01$ & $68.36$ & $79.16$ & $89.71$ &&
                                     $50.90$ & $60.67$ & $72.43$ & $85.74$ &&
                                     $65.5$ & $1.00\times$ \\
    $+$ \rev{frame attention in early blocks} & $56.25$ & $67.01$ & $78.35$ & $89.32$ &&
                                     $50.64$ & $60.39$ & $72.18$ & $85.63$ &&
                                     $35.4$ & $1.85\times$ \\
    $+$ \rev{pooled cross-frame broadcast} & $58.06$ & $68.42$ & $79.18$ & $89.66$ &&
                                     $51.34$ & $60.85$ & $72.42$ & $85.67$ &&
                                     $36.5$ & $1.80\times$ \\
    $+$ \rev{query-preserving K/V merging}, $r{=}0.50$ & $\mathbf{57.29}$ & $\mathbf{67.92}$
                                   & $\mathbf{78.97}$ & $\mathbf{89.56}$ &&
                                     $\mathbf{51.04}$ & $\mathbf{60.66}$
                                   & $\mathbf{72.35}$ & $\mathbf{85.73}$ &&
                                     $\mathbf{27.7}$ & $\mathbf{2.36\times}$ \\
    $+$ \rev{query-preserving K/V merging}, $r{=}0.90$ & $56.40$ & $67.34$ & $78.58$ & $89.35$ &&
                                     $49.85$ & $59.71$ & $71.72$ & $85.45$ &&
                                     $23.8$ & $2.75\times$ \\
    \bottomrule
  \end{tabular}
  \caption{\textbf{Ablation of the proposed design on $\pi^3$.} Bold
  marks the adopted configuration.}
  \label{tab:supp:transfer:pi3}
\end{table*}

\begin{table*}[!t]
  \centering\footnotesize
  \setlength{\tabcolsep}{4pt}
  \renewcommand{\arraystretch}{1.05}
  \begin{tabular}{@{}lrrrrcrrrrcrr@{}}
    \toprule
      & \multicolumn{4}{c}{CO3D AUC@$\tau$} && \multicolumn{4}{c}{RE10K AUC@$\tau$}
      && \multicolumn{2}{c}{$200$ frames} \\
    \cmidrule(lr){2-5}\cmidrule(lr){7-10}\cmidrule(lr){12-13}
      & $3$ & $5$ & $10$ & $30$ && $3$ & $5$ & $10$ & $30$
      && time (s) & speedup \\
    \midrule
    vanilla                        & $54.66$ & $66.09$ & $77.95$ & $89.14$ &&
                                     $63.22$ & $71.26$ & $80.59$ & $90.58$ &&
                                     $41.5$ & $1.00\times$ \\
    $+$ \rev{frame attention in early blocks} & $53.98$ & $65.59$ & $77.62$ & $88.97$ &&
                                     $63.01$ & $71.09$ & $80.47$ & $90.57$ &&
                                     $29.4$ & $1.41\times$ \\
    $+$ \rev{pooled cross-frame broadcast} & $54.56$ & $66.01$ & $77.90$ & $89.11$ &&
                                     $63.21$ & $71.23$ & $80.59$ & $90.61$ &&
                                     $29.9$ & $1.39\times$ \\
    $+$ \rev{query-preserving K/V merging}, $r{=}0.50$ & $\mathbf{53.84}$ & $\mathbf{65.44}$
                                   & $\mathbf{77.52}$ & $\mathbf{89.00}$ &&
                                     $\mathbf{62.95}$ & $\mathbf{71.05}$
                                   & $\mathbf{80.50}$ & $\mathbf{90.56}$ &&
                                     $\mathbf{22.5}$ & $\mathbf{1.84\times}$ \\
    $+$ \rev{query-preserving K/V merging}, $r{=}0.90$ & $52.97$ & $64.81$ & $77.08$ & $88.81$ &&
                                     $62.64$ & $70.82$ & $80.36$ & $90.52$ &&
                                     $19.2$ & $2.17\times$ \\
    \bottomrule
  \end{tabular}
  \caption{\textbf{Ablation of the proposed design on VGGT-$\Omega$.} Bold
  marks the adopted configuration.}
  \label{tab:supp:transfer:omega}
\end{table*}

%% file: fig/s2_core_transfer.tex
\begin{figure*}[t]
  \centering
  \includegraphics[width=\linewidth]{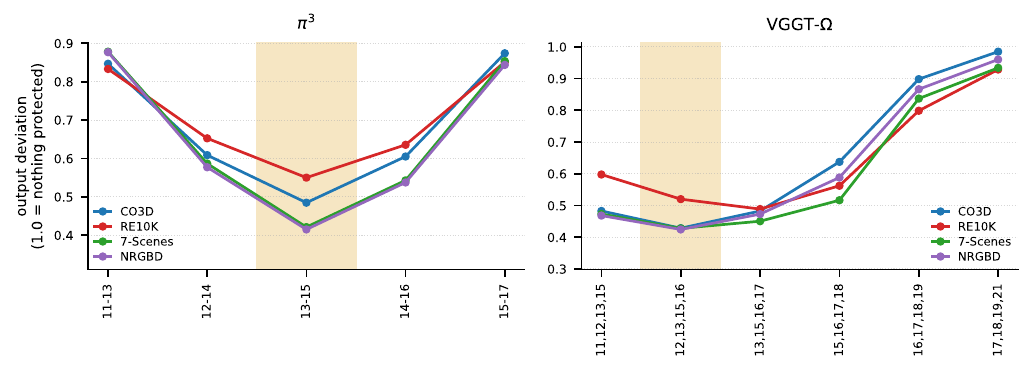}
  \caption{\textbf{Core-block selection on $\pi^3$ and VGGT-$\Omega$.} Core blocks
  are selected using the same measurement as in
  Figure~\ref{fig:supp:core}. The criterion selects blocks $13$--$15$ on $\pi^3$
  and raw blocks $12$, $13$, $15$, and $16$ on VGGT-$\Omega$, where raw block $14$
  is a register block and is therefore excluded from merging.}
  \label{fig:supp:coretransfer}
\end{figure*}

%% file: supp/I_qualitative.tex
\section{Additional Qualitative Results}
\label{sec:supp:qualitative}

\begin{figure*}[t]
  \centering
  \includegraphics[width=\linewidth]{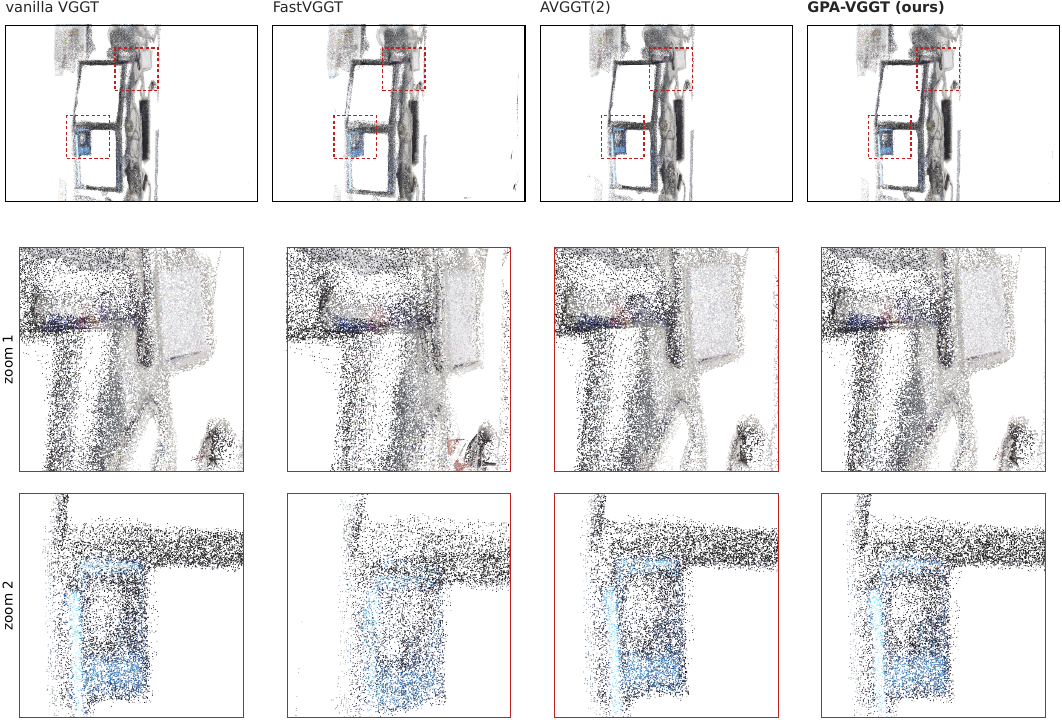}
  \caption{\textbf{A small-room scene from 7-Scenes} (\texttt{heads-seq-02\#0}).}
  \label{fig:supp:heads2b}
\end{figure*}

\begin{figure*}[t]
  \centering
  \includegraphics[width=\linewidth]{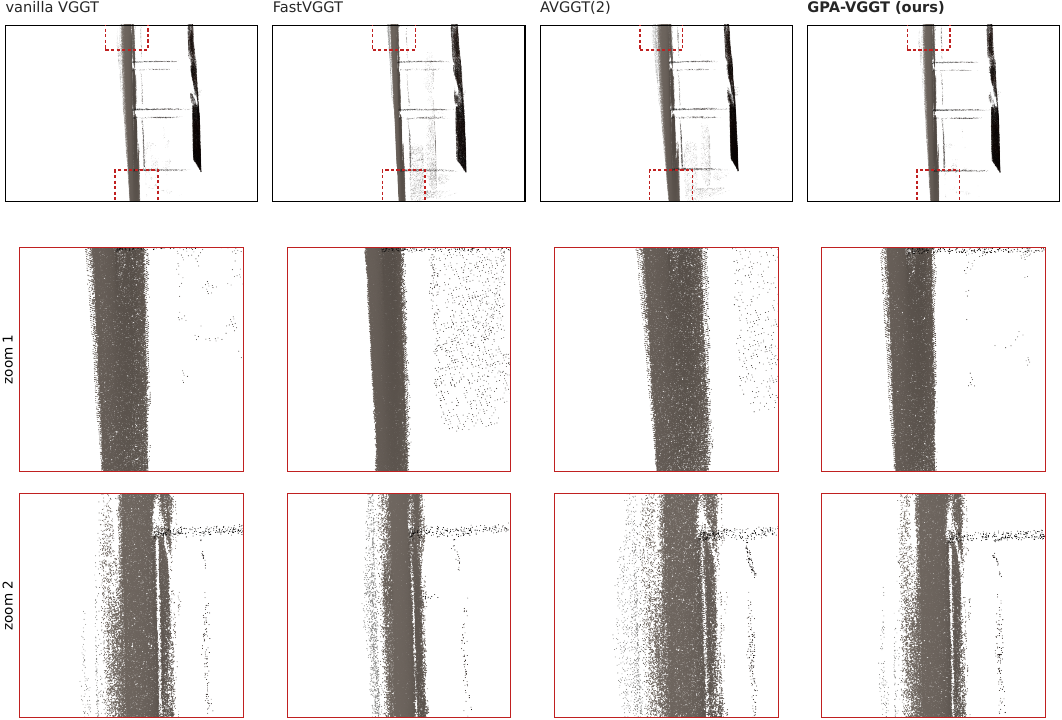}
  \caption{\textbf{A green-room scene from NRGBD} (\texttt{green\_room\#11}).}
  \label{fig:supp:green11}
\end{figure*}

\begin{figure*}[t]
  \centering
  \includegraphics[width=\linewidth]{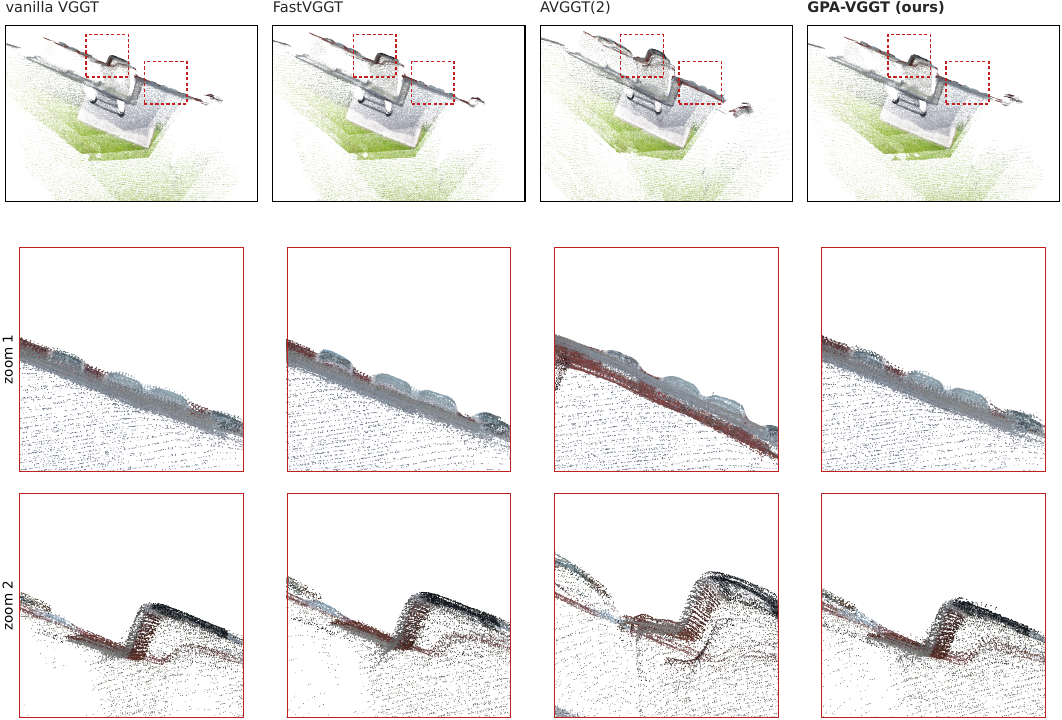}
  \caption{\textbf{An outdoor scene from RealEstate10K} (\texttt{2c805f56d92a2e22}).}
  \label{fig:supp:re10k}
\end{figure*}

Figures~\ref{fig:supp:heads2b}--\ref{fig:supp:re10k} provide additional
qualitative comparisons for examples from 7-Scenes, NRGBD, and RealEstate10K.

%% file: main.bbl
\begin{thebibliography}{25}
\providecommand{\natexlab}[1]{#1}
\providecommand{\url}[1]{\texttt{#1}}
\expandafter\ifx\csname urlstyle\endcsname\relax
  \providecommand{\doi}[1]{doi: #1}\else
  \providecommand{\doi}{doi: \begingroup \urlstyle{rm}\Url}\fi

\bibitem[Wang et~al.(2025)Wang, Chen, Karaev, Vedaldi, Rupprecht, and Novotny]{vggt}
Jianyuan Wang, Minghao Chen, Nikita Karaev, Andrea Vedaldi, Christian Rupprecht, and David Novotny.
\newblock {VGGT}: Visual geometry grounded transformer.
\newblock In \emph{CVPR}, 2025.

\bibitem[Shen et~al.(2026)Shen, Zhang, Qu, Zheng, Ji, Zhang, and Cao]{fastvggt}
You Shen, Zhipeng Zhang, Yansong Qu, Xiawu Zheng, Jiayi Ji, Shengchuan Zhang, and Liujuan Cao.
\newblock {FastVGGT}: Training-free acceleration of visual geometry transformer.
\newblock In \emph{ICLR}, 2026.

\bibitem[Sun et~al.(2026)Sun, Zhu, Lou, Yang, Tang, Zhang, Wang, and Zhang]{avggt}
Xianbing Sun, Zhikai Zhu, Zhengyu Lou, Bo Yang, Jinyang Tang, Liqing Zhang, He Wang, and Jianfu Zhang.
\newblock {AVGGT}: Rethinking global attention for accelerating {VGGT}.
\newblock In \emph{CVPR}, 2026.

\bibitem[Shu et~al.(2026)Shu, Lin, Xie, Yin, Li, Pu, Li, Yao, Cao, Guo, and Long]{litevggt}
Zhijian Shu, Cheng Lin, Tao Xie, Wei Yin, Ben Li, Zhiyuan Pu, Weize Li, Yao Yao, Xun Cao, Xiaoyang Guo, and Xiao-Xiao Long.
\newblock {LiteVGGT}: Boosting vanilla {VGGT} via geometry-aware cached token merging.
\newblock In \emph{CVPR}, 2026.

\bibitem[Wang and Xu(2026)]{flashvggt}
Zipeng Wang and Dan Xu.
\newblock {FlashVGGT}: Efficient and scalable visual geometry transformers with compressed descriptor attention.
\newblock In \emph{CVPR}, 2026.

\bibitem[Wang et~al.(2026)Wang, Schmidt, Piekenbrinck, and Leibe]{fastervggt}
Chung-Shien~Brian Wang, Christian Schmidt, Jens Piekenbrinck, and Bastian Leibe.
\newblock Block-sparse global attention for efficient multi-view geometry transformers.
\newblock In \emph{CVPR}, 2026.

\bibitem[Feng et~al.(2026)Feng, Qin, Wu, Yang, Li, Li, An, Huang, Zhang, Magno, and Xu]{quantvggt}
Weilun Feng, Haotong Qin, Mingqiang Wu, Chuanguang Yang, Yuqi Li, Xiangqi Li, Zhulin An, Libo Huang, Yulun Zhang, Michele Magno, and Yongjun Xu.
\newblock Quantized visual geometry grounded transformer.
\newblock In \emph{ICLR}, 2026.

\bibitem[Pan et~al.(2026)Pan, Wang, and Wang]{qvggt}
Zhizhen Pan, Hesong Wang, and Huan Wang.
\newblock {QVGGT}: Post-training quantized visual geometry grounded transformer.
\newblock In \emph{CVPR}, 2026.

\bibitem[Chen et~al.(2026)Chen, Qiu, Li, Agha, Omidshafiei, Patrikar, and Scherer]{come}
Yutian Chen, Yuheng Qiu, Ruogu Li, Ali Agha, Shayegan Omidshafiei, Jay Patrikar, and Sebastian Scherer.
\newblock {Co-Me}: Confidence-guided token merging for visual geometric transformers.
\newblock In \emph{CVPR}, 2026.

\bibitem[Wang et~al.(2026)Wang, Meiner, Rai, De~La~Parra, and Kumar]{httm}
Weitian Wang, Lukas Meiner, Shubham Rai, Cecilia De~La~Parra, and Akash Kumar.
\newblock {HTTM}: Head-wise temporal token merging for faster {VGGT}.
\newblock In \emph{CVPR}, 2026.

\bibitem[Wang et~al.(2024)Wang, Leroy, Cabon, Chidlovskii, and Revaud]{dust3r}
Shuzhe Wang, Vincent Leroy, Yohann Cabon, Boris Chidlovskii, and Jerome Revaud.
\newblock {DUSt3R}: Geometric {3D} vision made easy.
\newblock In \emph{CVPR}, 2024.

\bibitem[Yang et~al.(2025)Yang, Sax, Liang, Henaff, Tang, Cao, Chai, Meier, and Feiszli]{fast3r}
Jianing Yang, Alexander Sax, Kevin~J. Liang, Mikael Henaff, Hao Tang, Ang Cao, Joyce Chai, Franziska Meier, and Matt Feiszli.
\newblock {Fast3R}: Towards {3D} reconstruction of 1000+ images in one forward pass.
\newblock In \emph{CVPR}, 2025.

\bibitem[Wang et~al.(2026{\natexlab{a}})Wang, Zhou, Zhu, Chang, Zhou, Li, Chen, Pang, Shen, and He]{pi3}
Yifan Wang, Jianjun Zhou, Haoyi Zhu, Wenzheng Chang, Yang Zhou, Zizun Li, Junyi Chen, Jiangmiao Pang, Chunhua Shen, and Tong He.
\newblock $\pi^3$: Permutation-equivariant visual geometry learning.
\newblock In \emph{ICLR}, 2026{\natexlab{a}}.

\bibitem[Wang et~al.(2026{\natexlab{b}})Wang, Chen, Zhang, Karaev, Sch\"onberger, Labatut, Bojanowski, Novotny, Vedaldi, and Rupprecht]{vggtomega}
Jianyuan Wang, Minghao Chen, Shangzhan Zhang, Nikita Karaev, Johannes Sch\"onberger, Patrick Labatut, Piotr Bojanowski, David Novotny, Andrea Vedaldi, and Christian Rupprecht.
\newblock {VGGT}-{$\Omega$}.
\newblock In \emph{CVPR}, 2026{\natexlab{b}}.

\bibitem[Bolya et~al.(2023)Bolya, Fu, Dai, Zhang, Feichtenhofer, and Hoffman]{tome}
Daniel Bolya, Cheng-Yang Fu, Xiaoliang Dai, Peizhao Zhang, Christoph Feichtenhofer, and Judy Hoffman.
\newblock Token merging: Your {ViT} but faster.
\newblock In \emph{ICLR}, 2023.

\bibitem[Bolya and Hoffman(2023)]{tomesd}
Daniel Bolya and Judy Hoffman.
\newblock Token merging for fast stable diffusion.
\newblock In \emph{CVPRW}, 2023.

\bibitem[Reizenstein et~al.(2021)Reizenstein, Shapovalov, Henzler, Sbordone, Labatut, and Novotny]{co3d}
Jeremy Reizenstein, Roman Shapovalov, Philipp Henzler, Luca Sbordone, Patrick Labatut, and David Novotny.
\newblock Common objects in {3D}: Large-scale learning and evaluation of real-life {3D} category reconstruction.
\newblock In \emph{ICCV}, 2021.

\bibitem[Zhou et~al.(2018)Zhou, Tucker, Flynn, Fyffe, and Snavely]{re10k}
Tinghui Zhou, Richard Tucker, John Flynn, Graham Fyffe, and Noah Snavely.
\newblock Stereo magnification: Learning view synthesis using multiplane images.
\newblock \emph{ACM TOG}, 37\penalty0 (4), 2018.

\bibitem[Shotton et~al.(2013)Shotton, Glocker, Zach, Izadi, Criminisi, and Fitzgibbon]{sevenscenes}
Jamie Shotton, Ben Glocker, Christopher Zach, Shahram Izadi, Antonio Criminisi, and Andrew Fitzgibbon.
\newblock Scene coordinate regression forests for camera relocalization in {RGB-D} images.
\newblock In \emph{CVPR}, 2013.

\bibitem[Azinovi\'c et~al.(2022)Azinovi\'c, Martin-Brualla, Goldman, Nie{\ss}ner, and Thies]{nrgbd}
Dejan Azinovi\'c, Ricardo Martin-Brualla, Dan~B. Goldman, Matthias Nie{\ss}ner, and Justus Thies.
\newblock Neural {RGB-D} surface reconstruction.
\newblock In \emph{CVPR}, 2022.

\bibitem[Dao et~al.(2022)Dao, Fu, Ermon, Rudra, and R\'e]{flashattention}
Tri Dao, Daniel~Y. Fu, Stefano Ermon, Atri Rudra, and Christopher R\'e.
\newblock {FlashAttention}: Fast and memory-efficient exact attention with {IO}-awareness.
\newblock In \emph{NeurIPS}, 2022.

\bibitem[Tillet et~al.(2019)Tillet, Kung, and Cox]{triton}
Philippe Tillet, H.~T. Kung, and David Cox.
\newblock Triton: An intermediate language and compiler for tiled neural network computations.
\newblock In \emph{ACM SIGPLAN International Workshop on Machine Learning and Programming Languages (MAPL)}, 2019.

\bibitem[Paszke et~al.(2019)Paszke, Gross, Massa, Lerer, Bradbury, Chanan, Killeen, Lin, Gimelshein, Antiga, Desmaison, K\"opf, Yang, DeVito, Raison, Tejani, Chilamkurthy, Steiner, Fang, Bai, and Chintala]{pytorch}
Adam Paszke, Sam Gross, Francisco Massa, Adam Lerer, James Bradbury, Gregory Chanan, Trevor Killeen, Zeming Lin, Natalia Gimelshein, Luca Antiga, Alban Desmaison, Andreas K\"opf, Edward Yang, Zach DeVito, Martin Raison, Alykhan Tejani, Sasank Chilamkurthy, Benoit Steiner, Lu Fang, Junjie Bai, and Soumith Chintala.
\newblock {PyTorch}: An imperative style, high-performance deep learning library.
\newblock In \emph{NeurIPS}, 2019.

\bibitem[Oquab et~al.(2024)Oquab, Darcet, Moutakanni, Vo, Szafraniec, Khalidov, Fernandez, Haziza, Massa, El-Nouby, Assran, Ballas, Galuba, Howes, Huang, Li, Misra, Rabbat, Sharma, Synnaeve, Xu, Jegou, Mairal, Labatut, Joulin, and Bojanowski]{dinov2}
Maxime Oquab, Timoth\'{e}e Darcet, Th\'{e}o Moutakanni, Huy Vo, Marc Szafraniec, Vasil Khalidov, Pierre Fernandez, Daniel Haziza, Francisco Massa, Alaaeldin El-Nouby, Mahmoud Assran, Nicolas Ballas, Wojciech Galuba, Russell Howes, Po-Yao Huang, Shang-Wen Li, Ishan Misra, Michael Rabbat, Vasu Sharma, Gabriel Synnaeve, Hu Xu, Herv\'{e} Jegou, Julien Mairal, Patrick Labatut, Armand Joulin, and Piotr Bojanowski.
\newblock {DINOv2}: Learning robust visual features without supervision.
\newblock \emph{TMLR}, 2024.

\bibitem[Vaswani et~al.(2017)Vaswani, Shazeer, Parmar, Uszkoreit, Jones, Gomez, Kaiser, and Polosukhin]{transformer}
Ashish Vaswani, Noam Shazeer, Niki Parmar, Jakob Uszkoreit, Llion Jones, Aidan~N. Gomez, Lukasz Kaiser, and Illia Polosukhin.
\newblock Attention is all you need.
\newblock In \emph{NeurIPS}, 2017.

\end{thebibliography}
